\documentclass{article}
\usepackage{arxiv}
\usepackage[utf8]{inputenc}
\usepackage[T1]{fontenc}
\usepackage{hyperref}
\usepackage{url}
\usepackage{booktabs}
\usepackage{amsmath,amssymb,amsfonts}
\usepackage{microtype}
\usepackage{graphicx}
\usepackage{natbib}
\usepackage{placeins}

\usepackage{xcolor}
\newif\ifdraftnotes
\draftnotestrue

\newcommand{\figuretemplate}[2]{
  \IfFileExists{#1}{\includegraphics[width=\linewidth]{#1}}{
    \fbox{\begin{minipage}[c][#2][c]{0.94\linewidth}\centering
      \mbox{}
    \end{minipage}}}}
\title{CADFather: Autonomous CAD Reconstruction through Coordinated Tool Use}
\author{
\mdseries Gennadiy Savrasov\textsuperscript{1,4}
\and Maksim Elistratov\textsuperscript{1}
\and Nikita Gavrilov\textsuperscript{1}
\and Albert Garifullin\textsuperscript{1}
\AND
\mdseries Oleg Pavlov\textsuperscript{1,4}
\and Soslan Kabisov\textsuperscript{1}
\and Vladimir Frolov\textsuperscript{1}
\and Anton Konushin\textsuperscript{1,4}
\AND
\mdseries Andrey Kuznetsov\textsuperscript{2,3}
\and Dmitrii Zhemchuzhnikov\textsuperscript{1,4}
\AND
\mdseries
\textsuperscript{1} Lomonosov Moscow State University; \\
\textsuperscript{2} Innopolis University;
\textsuperscript{3} FusionBrain Lab; \\
\textsuperscript{4} AXXX, Moscow, Russia
}
\date{}
\renewcommand{\shorttitle}{CADFather}
\hypersetup{
  pdftitle={CADFather: Autonomous CAD Reconstruction through Coordinated Tool Use},
  pdfauthor={Gennadiy Savrasov, Maksim Elistratov, Nikita Gavrilov, Albert Garifullin, Oleg Pavlov, Soslan Kabisov, Vladimir Frolov, Anton Konushin, Andrey Kuznetsov, Dmitrii Zhemchuzhnikov},
  pdfkeywords={CAD reconstruction, CAD agents, vision-language models, tool-using agents}
}

\begin{document}
\maketitle

\begin{abstract}
Reconstructing an editable CAD model from a 3D shape remains a challenging engineering task. Existing methods can propose CAD operations, but no single source of proposals works equally well across different part geometries and stages of reconstruction. We introduce CADFather, an autonomous agentic system that coordinates complementary tools to recover parametric CAD programs from 3D meshes. A vision-language assistant inspects renders of the target and intermediate reconstructions, then decides which candidate CAD programs to extend, which tools to invoke, how many proposals to generate, and when to finish. Sampled stepwise generation and deterministic geometric methods propose CAD operations, while numerical optimization refines the parameters of existing programs. Proposed or refined programs are executed and evaluated to provide feedback for subsequent decisions. The agent maintains alternative candidate programs for each target part and preserves the best valid result throughout reconstruction. CADFather uses pretrained generation and assistant models without additional training. We evaluate reconstruction quality and execution validity on the full DeepCAD, Fusion360, and MCB test sets, as well as on CADENA-Bench, CADBench, and BenchCAD. We additionally analyze computational cost and the trade-off between cost and reconstruction quality.
\end{abstract}

\section{Introduction}
A mesh describes a shape, but it does not by itself encode the modeling operations and feature parameters used to construct that shape. Recovering an executable CAD program makes these quantities available for inspection and modification. For example, a suitable construction program can expose a bore diameter or wall thickness as an editable parameter. This capability is valuable when a component is available only as a scan, a legacy mesh, or a model without its construction history. The reconstruction must be executable, structurally coherent, and useful as an editable
model.

Recent work has explored several approaches to CAD reconstruction.. Some methods adopt a learned approach, fine-tuning vision-language models to predict CAD code from an input mesh, as in CAD-Recode, cadrille, CADEvolve, CADReasoner, and CADENA~\citep{cadrecode,cadrille,cadevolve,cadreasoner,cadena}. Within this class, CAD-Recode, cadrille and CADEvolve generate complete programs in a single pass. CADReasoner and CADENA instead reconstruct iteratively: CADReasoner revises complete programs using feedback on discrepancies with the target, while CADENA extends a partial program one operation at a time using renders of the target and current reconstruction. CADFit follows an algorithmic approach, reconstructing the target through geometric fitting and validation of candidate operations within an optimization-based search~\citep{cadfit}. Numerical parameter optimization adjusts an existing program while retaining its operation structure~\citep{differentiablecad,diffcsg}.

These approaches offer complementary ways to improve a reconstruction. A learned model can propose plausible modeling operations, but its proposals may miss a feature or use inaccurate dimensions. Algorithmic construction provides an additional source of operations derived directly from the target geometry, while parameter optimization can refine the dimensions of an otherwise suitable program. Combining these capabilities allows the system to explore alternative constructions and improve the fit of existing ones. The useful action depends on the current reconstruction: it may need another operation, parameter refinement, or a return to an earlier candidate. This motivates a process that selects both the candidate to develop and the tool to apply using feedback from the executed geometry.

We introduce \emph{CADFather}, which maintains a pool of alternative CAD programs and uses a vision-language assistant to guide their reconstruction. At each step, the assistant inspects target and candidate renders, geometric scores, available actions, and the history of executed requests. It then selects candidates to extend or refine, chooses tools, specifies how many learned proposals to sample, or ends reconstruction. The tools comprise \emph{learned proposal generation}, \emph{algorithmic proposal generation}, and \emph{parameter optimization}. Learned proposals come from a pretrained operation generator; algorithmic proposals are constructed from geometry without a learned models. Parameter optimization refines an existing program. Each proposed program is executed and measured. The assistant then inspects its rendered geometry and decides whether to accept it into the candidate pool. Earlier candidates remain available, and the system preserves the best valid result throughout the search.

CAD-Assistant and IterCAD already show how visual agents can use executable CAD environments and inspect the outcomes of their actions~\citep{cadassistant,itercad}. CADFather applies this interaction pattern to a pool of alternative mesh reconstructions, with specialized proposal and refinement tools. Our focus is the integration of these capabilities in a shared representation and search process. Neither the operation generator nor the assistant requires additional training.

Our contributions are:
\begin{itemize}
\item a mesh-to-CAD reconstruction agent that uses visual and execution feedback to select candidate programs, choose reconstruction tools, and allocate learned proposal samples;
\item a shared reconstruction framework combining learned proposals, algorithmic proposals, parameter optimization, and persistent retention of the best valid result;
\item an evaluation on six datasets, with published-method comparisons and ablations removing algorithmic proposals, parameter optimization, or both.
\end{itemize}
The experiments assess geometric fidelity, validity, and selected measures of computational cost. They evaluate the complete system and the contribution of making the different tools available during reconstruction.

\section{Related Work}
\paragraph{Generative CAD modeling.}
Many learning-based CAD methods generate sequences of construction operations and their parameters, rather than only meshes or point clouds. DeepCAD learns distributions over sketch--extrude construction sequences~\citep{deepcad}, while the Fusion360 Gallery provides human modeling histories and an environment for programmatic construction~\citep{fusion360}. SkexGen separates aspects of topology and geometry through disentangled codebooks for construction-sequence generation~\citep{skexgen}. A complementary line models the sketches underlying these operations: SketchGen and Vitruvion generate parametric sketches with constraints~\citep{sketchgen,vitruvion}, while CadVLM incorporates visual and language inputs for sketch generation~\citep{cadvlm}. These methods establish representations and generative models for CAD construction, but generating a plausible design is distinct from recovering a program for a specified target mesh. CADFather targets the same reconstruction task but uses existing CAD modeling capabilities within an agent to improve reconstruction quality.

\paragraph{Geometric reconstruction and executable CAD programs.}
CAD reconstruction methods differ in the representation they recover. ComplexGen reconstructs boundary representations through their geometric and topological structure~\citep{complexgen}; SolidGen and BrepGen study direct generation of boundary representations~\citep{solidgen,brepgen}. Constructive solid geometry methods such as CSGNet, UCSG-Net, and CAPRI-Net describe shapes through primitive combinations~\citep{csgnet,ucsgnet,caprinet}. Point2Cyl recovers extrusion cylinders from point clouds, while ExtrudeNet and SECAD-Net learn inverse sketch-and-extrude representations~\citep{point2cyl,extrudenet,secadnet}. These approaches recover structured geometry, although their outputs need not provide general-purpose CAD code with an explicit modeling sequence. Methods targeting executable programs include CAD-Recode, which maps point clouds to CadQuery code~\citep{cadrecode}, and cadrille, which supports multimodal reconstruction and uses reinforcement learning with geometric rewards~\citep{cadrille}. CAD-Coder generates CAD code from visual inputs~\citep{cadcodervlm}, while CADEvolve develops training data through CAD program evolution~\citep{cadevolve}. CADFather targets executable construction programs and combines learning-based operation generation with geometry-based operation proposals through a common tool interface.

\paragraph{Iterative and stepwise CAD reconstruction.}
Executing intermediate programs makes their resulting geometry available to guide subsequent reconstruction. Zone-graph methods search over CAD modeling sequences~\citep{zonegraphs}, and CADFit combines incremental operation construction with hybrid optimization against the target mesh~\citep{cadfit}. CADReasoner uses geometric discrepancies to iteratively edit complete CAD programs~\citep{cadreasoner}. CADENA adopts a different prediction unit: it executes a partial program, renders its geometry together with the target, and generates one additional operation~\citep{cadena}. SOV-CAD also investigates stepwise reconstruction with orthographic visual feedback, using a Decision Transformer formulation~\citep{sovcad}. These approaches differ in what they predict and how they use intermediate geometry: some revise complete programs, while others construct them operation by operation. CADFather builds on the latter setting but makes tool selection a separate decision of the visual assistant. At each decision step, it chooses which candidate CAD program to continue, whether to request stepwise or deterministic proposals and how many, or whether the current candidate is ready for parameter refinement or completion.

\paragraph{Parameter optimization in CAD programs.}
Numerical refinement can adjust the geometry represented by a program without changing its operation structure. \citet{differentiablecad} introduce a differentiable CAD language and formulate direct geometric edits as constrained optimization of program parameters. Their system seeks parameter updates that satisfy editing objectives while preserving program validity, rather than reconstructing an unknown modeling sequence. DiffCSG~\citep{diffcsg} enables differentiable rasterization of constructive solid geometry, supporting optimization of primitive parameters for direct and image-based editing with a prescribed discrete structure. These works distinguish continuous parameter refinement from the selection of modeling operations. CADFather exposes parameter optimization as a separate tool: the assistant decides when an existing candidate should be refined rather than extended or replaced by an alternative proposal. We do not introduce a new numerical optimizer.

\paragraph{Tool-using CAD agents.}
Agentic systems make the selection and execution of modeling actions explicit. CAD-Assistant uses a vision-language planner with CAD-specific tools and a FreeCAD execution environment, adapting its actions to the evolving design state~\citep{cadassistant}. Its tools include rendering and cross-section extraction, allowing the planner to obtain additional observations rather than relying only on the initial input. IterCAD formulates visually grounded generation and editing as multimodal interaction with an executable sandbox~\citep{itercad}. ToolCAD studies reinforcement learning for tool-using language models in text-to-CAD generation~\citep{toolcad}, while COSMO-Agent investigates tool orchestration across modeling, simulation, and optimization~\citep{cosmoagent}. CAD-Assistant already establishes vision-language planning with CAD tools and execution feedback; tool use alone is not our distinction. CADFather focuses on reconstructing an executable program for a given target mesh while maintaining alternative candidate CAD programs for that part. Its visual assistant uses intermediate renders and geometric evaluation to choose both a candidate to continue and a specialized tool to apply. This couples target-conditioned tool choice with search over alternative reconstruction branches, rather than relying on a fixed operation source or repeatedly editing a single program.

\section{Method}
\label{sec:method}
\subsection{Problem Formulation and System Overview}
Given a target mesh $M^\star$, CADFather seeks an executable CAD program $p$ whose resulting geometry $\mathcal{E}(p)$ matches the target. For learned stepwise prediction of CAD operations, we use CADENA-RL~\citep{cadena}. Candidate programs therefore follow CADENA's CadQuery-based DSL, in which each line specifies a modeling operation that produces a buildable and renderable solid. All tools produce programs in this representation, so a candidate generated by one tool can subsequently be extended or refined using another. For each target part, the system maintains alternative candidate programs $p_i$ and their evaluation results. Successfully executed programs provide valid states for further reconstruction; execution failures are recorded but cannot be selected for extension.

\begin{figure}[!tbp]
\centering
\figuretemplate{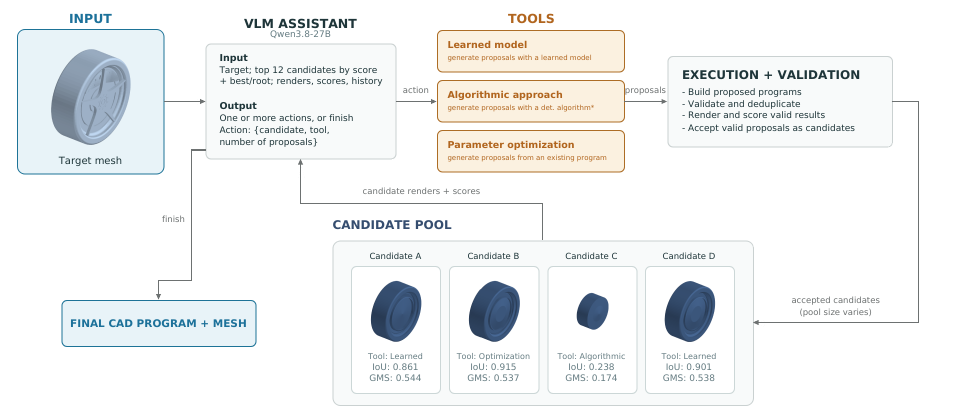}{4.2cm}
\caption{CADFather overview. Given a target mesh and feedback from existing CAD candidates, the visual assistant selects candidate programs to continue, tools to apply, and the number of proposals to generate. Stepwise generation and deterministic proposals create alternative continuations, while parameter optimization refines existing candidates. The resulting programs are executed and evaluated; their renders and scores inform the assistant's next decision step.
\textsuperscript{*} ``det.'' abbreviates ``deterministic.''}
\label{fig:overview}
\end{figure}

At each decision step, the vision-language assistant inspects the target and the current candidates, then chooses which candidate programs to continue, which generation or refinement tools to invoke, how many proposals to request, or whether to finish. The generation tools, sampled stepwise generation and deterministic CAD proposals, offer alternative next operations, while the refinement tool, parameter optimization, adjusts an existing program without adding an operation. The agent's execution component runs the resulting programs, validates their geometry, evaluates candidates, manages the search budget, and preserves the best valid result independently of the assistant's current choices. The generation and assistant models require no additional training. Figure~\ref{fig:overview} summarizes this loop.

\subsection{Candidate Pool}
\label{sec:pool}
Each candidate records its identifier, parent, source tool, program, generated mesh, number of CAD operations, validity, and measurements. The root represents an empty program. Extending a candidate creates a child rather than modifying its stored history. Numerical optimization may change existing parameters and is therefore distinguished from an appended CAD operation.

The pool retains candidates from earlier decision steps. The assistant is not restricted to the latest generation: it can revisit an earlier valid CAD program or request a different first operation from the root, provided the corresponding actions are available. Code-level deduplication prevents an identical program from creating a new pool entry. This is textual deduplication, not a guarantee of geometric uniqueness.

At each decision step, the assistant receives a bounded table summarizing selected candidates from the pool, rather than the entire pool. Only candidates that built, are watertight, and still admit a legal action are listed. Up to twelve candidates are selected in descending order of score (Eq.~\ref{eq:score}). The protected best and the empty start are also shown if they are not among those selected. Each row gives the candidate identifier, the tool that produced it, its operation count, a status mark (start or best), its score with the separate volume and surface metrics (IoU and GMS, Section~\ref{sec:measure}), the score of the first operation of its chain, and the tools that may still be applied to it. The first-operation score helps the assistant judge whether to continue a branch or request an alternative first operation from the empty program. A low initial score is a warning signal, not a bound on the quality that later operations can achieve. A tool that is exhausted on a candidate, or does not apply to it, disappears from that row; a candidate absent from the table cannot be referenced in that step.

\subsection{Assistant Decision Step}
\label{sec:dialogue}

\begin{figure}[!tbp]
\centering
\figuretemplate{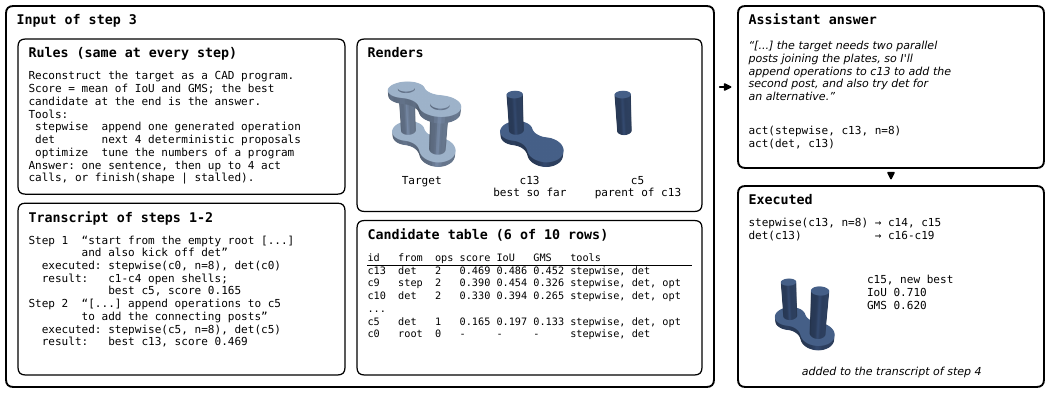}{5.0cm}
\caption{One assistant decision step: step 3 of the trajectory in Fig.~\ref{fig:trajectory} (MCB chain link). The best candidate c13 has one plate and one post, while the target has two of each. The assistant continues c13 with stepwise generation and deterministic proposals; stepwise generation adds the second post (c15), and the results of both calls enter the transcript of the next step. Abridged from the logged prompt and answer: the rules are paraphrased, the transcript and the rationale are excerpts, the table shows 6 of 10 rows, det denotes deterministic proposals, and one of four rendered views is shown.}
\label{fig:assistant_step}
\end{figure}

A decision step is made for the whole part, not for each candidate: in one step the assistant sees the entire candidate table and may act on several candidates at once. Its input has four parts. The first is a rule description that is the same for every step and every part: the task, the objective, the available tools with their approximate cost, and the answer format. The second is a transcript of the reconstruction so far. For every previous step, it records the assistant's one-sentence rationale and the actions the agent actually executed: requests for unavailable actions are dropped, and sample counts are reduced to fit the remaining budget. Recording executed rather than requested actions keeps the transcript consistent with the candidates that actually exist in the pool. The third part is a set of renders, described below, and the fourth is the candidate table (Section~\ref{sec:pool}). From this input the assistant decides whether to sample further operations with the learned generation model, request deterministic proposals, refine an existing program, or finish the part. Figure~\ref{fig:assistant_step} shows one complete step.
 
The renders do not show every candidate, only the target and the current state of the search. The render of the target is the same on every step. Once a candidate has been built, the prompt also shows two candidates: the best one so far and the best one produced by the previous step. If the previous step itself produced the new best, the second image shows its parent instead, so that the assistant sees the effect of the last operation; if the previous step produced no valid candidate, only the best is shown. Each image is labeled with the id from the candidate table, its operation count, and its score. Renders from earlier steps are not kept in the prompt.

The assistant answers through function calls. It first writes one short sentence stating what it pursues in this step; this sentence, unlike any further reasoning, is carried into the transcript. It then issues up to four \texttt{act} calls, each naming a tool and a candidate from the table, and for stepwise generation a sample count $n$. A \texttt{finish} call ends the whole part with one of two reasons (Section~\ref{sec:termination}); it may accompany \texttt{act} calls, in which case the step is the last one.

\subsection{Reconstruction Tools}
\label{sec:tools}
\paragraph{Sampled stepwise generation.}
For a parent program $p_i$, the stepwise generation tool produces $n$ next-operation proposals in a single request. Each accepted proposal forms a child program by continuing the selected parent's program. The count $n$ denotes generated alternatives, not $n$ sequential operations within one child. The assistant chooses $n$ up to a limit printed in the candidate table: the smaller of a fixed per-call cap and the generator samples still left in the part's budget. The limit applies to each call; when several stepwise calls in one step together exceed the remaining budget, the execution component reduces $n$ of the later ones or drops them.

\paragraph{Deterministic CAD operation proposals.}
The deterministic proposal tool builds CAD operations from the shape of the target, without the generation model. It follows the extrusion fitting of CADFit~\citep{cadfit}. The assistant decides when and on which candidate to call it. The tool has two entry points.

On the empty start, the tool proposes a first operation from the target mesh alone. If the part is a body of revolution, the first proposal revolves its profile. The tool then looks for extrusions: it cuts the target with planes, takes each cross-section as a sketch, and finds the depth over which this sketch follows the target surface. Extrusions that add little to the overlap with the target are dropped, so each remaining one matches a separate feature of the part.

On an existing candidate, the tool proposes the next operation; we call these warm proposals. It compares the candidate with the target and finds two regions: material that is missing and material that is extra. The same extrusion search runs on each region. Missing material gives extrusions to add, and extra material gives cuts.

The proposals form a ranked list. Unlike stepwise generation, the assistant does not choose how many proposals to take: each call returns the next four proposals from the list. Each entry is appended to the parent as a separate child, so four proposals give four alternative programs, not one program with four new operations. The children are executed and measured like any other candidate. A repeated request continues the same list, and when the list is exhausted, the tool disappears from that candidate's row. The tool matters most on parts where every first operation sampled from the generator gives an open surface rather than a solid. The search cannot continue from such a result, while an extrusion of a cross-section of the target gives a closed solid to build on (Section~\ref{sec:ablations}).

\paragraph{Parameter optimization.}
The optimizer changes the numbers in an existing candidate program but keeps its operations. The free parameters are the sketch coordinates and radii and the extrusion depths; the sketch planes and the revolve axes and angles stay fixed.

To compare the program with the target, the optimizer uses signed distances: for a point in space, the distance to the surface of a solid, negative inside it. The distance to the program's solid is computed directly from its operations, and the distance to the target is computed from the mesh at sampled points. The mismatch between the program and the target is the mean squared difference of the two distances over these points. Its gradient is estimated numerically: each parameter in turn is shifted by a small step, and the change in the mismatch gives the derivative with respect to that parameter. The optimizer then updates all parameters and repeats the step. It first fits points spread over the whole part, then points close to the target surface. Of the starting, coarse, and fine parameters, it keeps those that match best near the surface, so the program may also come back unchanged.

The refined program becomes a new child of the selected candidate with the same operations and is executed and measured like any other candidate. The optimizer handles extrusions, holes, and revolves; programs with other operations cannot be refined. It fits distances, not the IoU/GMS score, so whether the child is better is decided by the score (Eq.~\ref{eq:score}). The tool is deterministic and is therefore offered once per candidate.

\subsection{Execution, Measurements, and Protected Best}
\label{sec:measure}
A tool response is only a proposal: it becomes a candidate once the agent has executed the program and measured the result. Execution runs separately from the assistant. A candidate is invalid if its program does not build or its mesh is not, as a whole, a watertight solid with positive volume. This is stricter than the evaluation code of CADENA~\citep{cadena}, which accepts a prediction if any connected component is a closed solid and computes IoU on that component. A candidate whose execution or measurement exceeds the time limit also counts as invalid. The execution component records which of these happened. Whether a candidate is valid and how close it is to the target are reported separately.

Before measuring, the target and the candidate are placed in a common frame. The target is centered by its bounding box and scaled uniformly by its largest dimension. The candidate is brought to the same scale by one fixed transform of the modeling coordinates, shared by all candidates, rather than by fitting its own bounding box. A candidate of the wrong size therefore scores lower instead of having its scale error normalized away.

Two metrics compare a candidate with the target. Intersection-over-Union (IoU) measures volumetric overlap:
\begin{equation}
 I_i = \frac{V(M^\star \cap \mathcal{E}(p_i))}{V(M^\star \cup \mathcal{E}(p_i))}.
\end{equation}
The Generalized Match Score (GMS) measures how well the two surfaces and their normals agree. We use the definition of~\citet{cadena}, whose raw value is an area over a curve on the angle range $[0^\circ,25^\circ]$ and therefore lies in $[0,25]$, and normalize it as $H_i=1-\mathrm{GMS}_{\mathrm{raw},i}/25$, so that both metrics lie in $[0,1]$ and higher is better. IoU requires a target that is a single watertight solid with nonzero volume; for any other target only GMS is available.

The score of a candidate is the mean of the metrics available for it:
\begin{equation}
 s_i=\frac{1}{|\mathcal{Q}_i|}\sum_{q\in\mathcal{Q}_i}q,
 \qquad \mathcal{Q}_i\subseteq\{I_i,H_i\},
 \label{eq:score}
\end{equation}
that is, the average of IoU and GMS, or GMS alone for a target without IoU. An invalid candidate scores $s_i=0$.

The agent maintains the highest-scoring candidate as the \emph{protected best}. It is updated as soon as each candidate is evaluated, so a good candidate is kept even if the assistant never returns to it or continues another branch. A valid candidate always takes precedence over an invalid one. The final output is this protected best, not the last candidate the assistant worked on. The assistant may judge progress in its own terms when choosing tools, but only Eq.~\ref{eq:score} decides which candidate is returned. Since the target mesh is the input of the task, this selection uses no information beyond the input; it is, however, a best-of-$N$ choice by the same metrics that are reported in Section~\ref{sec:experiments}.

\subsection{Budgets, Recovery, and Termination}
\label{sec:termination}
Each part has a limited budget for every tool. The candidate table shows what is still available: a tool disappears from a row once its budget is spent, and the sample limit shrinks as generator samples are used. A request that exceeds the budget is reduced or rejected, and the assistant is told why.

A \texttt{finish} call ends the reconstruction of the \emph{whole part}, not of a single branch. The assistant gives one of two reasons: \texttt{shape}, when the best candidate already reproduces every feature of the target and only small numerical differences remain, or \texttt{stalled}, when some feature is still wrong but recent steps no longer improve the result and no other action seems worth trying. The prompt presents \texttt{stalled} as a normal outcome, not a failure. A \texttt{finish} call is ignored until at least one valid candidate exists. Independently of the assistant, fixed limits on the number of steps and on running time guarantee that every part terminates.

\begin{figure}[!tbp]
\centering
\figuretemplate{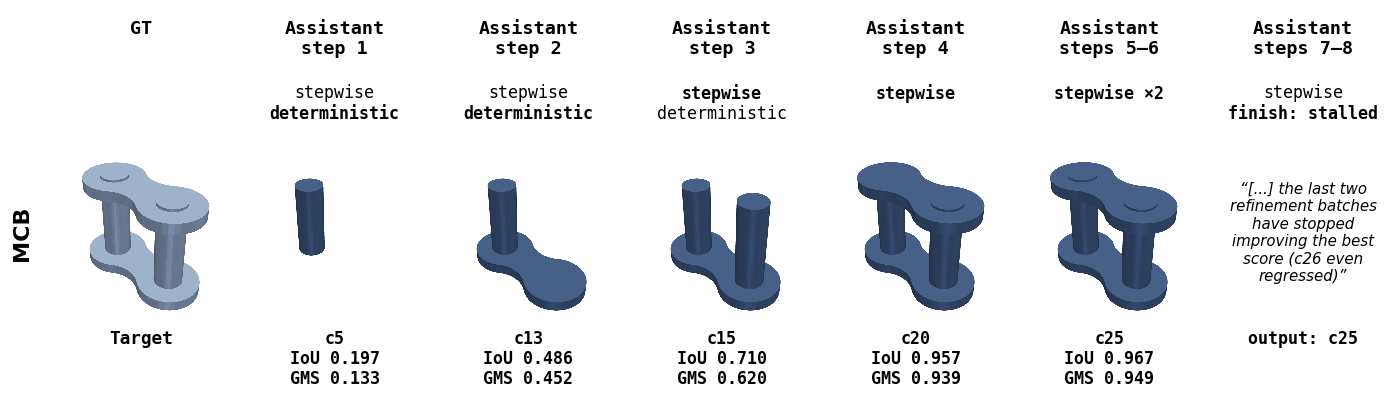}{4.0cm}
\caption{A reconstruction trajectory on an MCB part (a chain link). The figure follows the lineage of the final result, one column per assistant step or group of steps: above each render are the tools called in those steps, with the tool that produced the shown candidate in bold; below it are the candidate ID, IoU, and GMS. The deterministic start c5 recovers one post, a warm deterministic proposal adds the lower plate (c13), and stepwise operations add the second post (c15) and the upper plate (c20); in the same steps the stepwise first operations produce only open shells, and the other children score lower. In steps 5--6 the gains become small; the candidate c26 from step 7 has a higher IoU but a lower GMS than c25, and in step 8 the assistant ends the part with \texttt{finish(stalled)}, and the output is the protected best c25.}
\label{fig:trajectory}
\end{figure}

A malformed reply from the assistant does not end the part. If the reply calls no tool or requests only actions that are not available, the agent requests a revised action, stating the reason and the available actions. If this also fails, the agent performs the cheapest available action on the best candidate and tells the assistant about it. Failed executions are likewise reported back to the assistant, and the protected best is never lost. Figure~\ref{fig:trajectory} follows one reconstruction from the first step to the \texttt{finish} call.

\FloatBarrier

\section{Experiments}
\label{sec:experiments}
\subsection{Datasets and Setup}
We use the full test sets of DeepCAD, Fusion360, and MCB, containing 8046, 1725, and 5000 parts, respectively~\citep{deepcad,fusion360,mcb}. Ablations use fixed 1000-part subsets of each test set. In addition, we evaluate once on CADENA-Bench~\citep{cadena}, a benchmark of 3396 mechanical parts in six families drawn from TMCAD~\citep{tmcad}, MCB~\citep{mcb}, and CADNET~\citep{cadnet}. We also evaluate once, with the same configuration, on two recent benchmarks: CADBench~\citep{cadbench}, 18000 parts in six families of 3000 (Base, Fusion, Extrude, and All-Ops from STEP models, Mechanical and Organic from meshes), using its clean target meshes; and BenchCAD~\citep{benchcad}, 17900 CadQuery programs from 106 industrial part families in three difficulty tiers.

The CADENA-RL checkpoint used for stepwise operation proposals (Section~\ref{sec:method}) is served in bfloat16. The assistant is Qwen3.8-27B with FP8 weights; explicit thinking is disabled. Both models are frozen. Per part, the limits are 24 assistant steps, at most 12 operations in a program, and 1000 seconds of wall time; the budgets are 80 generator samples, 6 deterministic-proposal calls, 10 optimizer calls, and 120 program executions. A single stepwise call requests at most 32 samples. Building and measuring one candidate is limited to 30 seconds. The tool set is the one described in Section~\ref{sec:tools}.

\subsection{Metrics and Comparison Protocol}
\label{sec:protocol}

\begin{table}[!tbp]
 \centering
\caption{Comparison on the full test sets. All rows except CADFather are published results~\citep{cadena}: the baselines (CAD-Recode~\citep{cadrecode}, cadrille~\citep{cadrille}, CADReasoner~\citep{cadreasoner}, CADEvolve~\citep{cadevolve}, CADFit~\citep{cadfit}), CADENA-SFT and CADENA-RL greedy from its external-dataset table, CADENA-RL sampling from its decoding-strategy table ($T=1$, 12 samples per operation). Published CADENA rows use CADENA's own validity rule, which is more lenient than ours (Section~\ref{sec:measure}); their IR is not directly comparable to the IR of CADFather. All means are taken over valid predictions only, so a high IR leaves the means computed over an easier subset. $^\dagger$CADFit is an optimization loop run under a fixed compute budget; unconverged parts are counted in IR, and its prediction and target are normalized alike. IoU and GMS are means, CD values are medians scaled by $10^3$, and IR is a fraction.}
 \label{tab:comparison}
 \small
 \begin{tabular}{llrrrrr}
 \toprule
 Dataset & Method & IR & IoU & GMS & CD$_{8k}$ & CD$_{30k}$ \\
  \midrule
DeepCAD & CAD-Recode & 0.0790 & 0.914 & 0.929 & 0.17 & 0.055 \\
 & cadrille & 0.0030 & 0.897 & 0.948 & 0.17 & 0.059 \\
 & CADReasoner & 0.0300 & 0.910 & 0.949 & 0.16 & 0.049 \\
 & CADEvolve & 0.0090 & 0.924 & 0.953 & 0.16 & 0.050 \\
 & CADFit$^\dagger$ & 0.2750 & 0.914 & 0.898 & 0.18 & 0.052 \\
 & CADENA-SFT greedy & 0.0260 & 0.917 & 0.953 & 0.16 & 0.048 \\
 & CADENA-RL greedy & 0.0030 & 0.961 & 0.970 & \underline{0.15} & \underline{0.042} \\
  & CADENA-RL sampling & \underline{0.0007} & \underline{0.966} & \underline{0.974} & \underline{0.15} & \underline{0.042} \\
  & CADFather & \textbf{0.0000} & \textbf{0.987} & \textbf{0.983} & \textbf{0.138} & \textbf{0.039} \\
 \midrule
Fusion360 & CAD-Recode & 0.1360 & 0.872 & 0.858 & 0.17 & 0.060 \\
 & cadrille & 0.0080 & 0.848 & 0.868 & 0.17 & 0.066 \\
 & CADReasoner & 0.0720 & 0.863 & 0.886 & 0.15 & 0.050 \\
 & CADEvolve & 0.0270 & 0.881 & 0.882 & 0.16 & 0.056 \\
 & CADFit$^\dagger$ & 0.2560 & 0.882 & 0.846 & 0.17 & 0.054 \\
 & CADENA-SFT greedy & 0.0380 & 0.888 & 0.908 & 0.14 & 0.046 \\
 & CADENA-RL greedy & 0.0120 & 0.941 & 0.933 & \underline{0.12} & 0.038 \\
  & CADENA-RL sampling & \underline{0.0023} & \underline{0.952} & \underline{0.942} & \underline{0.12} & \underline{0.036} \\
  & CADFather & \textbf{0.0000} & \textbf{0.976} & \textbf{0.960} & \textbf{0.115} & \textbf{0.034} \\
 \midrule
MCB & CAD-Recode & 0.2150 & 0.711 & 0.547 & 0.66 & 0.524 \\
 & cadrille & 0.0150 & 0.662 & 0.550 & 0.85 & 0.720 \\
 & CADReasoner & 0.3170 & 0.690 & 0.552 & 1.44 & 1.290 \\
 & CADEvolve & 0.0920 & 0.729 & 0.583 & 0.63 & 0.500 \\
 & CADFit$^\dagger$ & 0.2970 & 0.758 & 0.601 & 0.48 & 0.365 \\
 & CADENA-SFT greedy & 0.1200 & 0.752 & 0.634 & 0.43 & 0.259 \\
 & CADENA-RL greedy & 0.0070 & 0.882 & 0.737 & \underline{0.22} & 0.093 \\
  & CADENA-RL sampling & \underline{0.0006} & \underline{0.895} & \underline{0.751} & \underline{0.22} & \underline{0.089} \\
  & CADFather & \textbf{0.0000} & \textbf{0.913} & \textbf{0.766} & \textbf{0.210} & \textbf{0.083} \\
\bottomrule
\end{tabular}
\end{table}

Besides IoU and GMS (Section~\ref{sec:measure}), Chamfer distance (CD) measures distances between sampled surfaces. We report the invalidity ratio (IR), mean IoU, mean GMS, and median CD at 8192 and 30000 sampled points, with CD multiplied by $10^3$. Validity, the common frame, and the GMS normalization are those of Section~\ref{sec:measure}; the frame and the GMS normalization coincide with the protocol of CADENA~\citep{cadena}, while our validity rule is stricter than CADENA's. IoU is computed only for targets whose mesh is a single watertight solid with nonzero volume; GMS and CD cover all targets, including open ones. Unless stated otherwise, IoU, GMS, and CD are taken over valid predictions, as in CADENA. Published CADENA results are taken from the original paper.

\subsection{Main Results}

\begin{table}[!tbp]
\centering
\caption{CADFather on CADENA-Bench by part family, and over all parts against published results. $n$ is the number of parts; IoU and GMS are means, CD values are medians over all parts scaled by $10^3$. The tooling-and-gauges family has 17 parts and is shown for completeness only. Published rows come from the full-benchmark table of~\citet{cadena}; there CD, IoU, and GMS are computed over valid predictions only, so each row describes a different subset whose size is given by IR. $^\dagger$As in Table~\ref{tab:comparison}.}
\label{tab:cadenabench}
\small
\begin{tabular}{lrrrrrr}
\toprule
Family & $n$ & IR & IoU & GMS & CD$_{8k}$ & CD$_{30k}$ \\
\midrule
Shafts and bushings & 762 & 0.0000 & 0.943 & 0.764 & 0.269 & 0.093 \\
Gears and bearings & 749 & 0.0000 & 0.880 & 0.660 & 0.423 & 0.241 \\
Housings and frames & 359 & 0.0000 & 0.917 & 0.772 & 0.385 & 0.245 \\
Levers and profiles & 463 & 0.0000 & 0.934 & 0.819 & 0.139 & 0.048 \\
Springs and fasteners & 1046 & 0.0000 & 0.890 & 0.672 & 0.320 & 0.173 \\
 Tooling and gauges & 17 & 0.0000 & 0.928 & 0.628 & 0.351 & 0.158 \\
 \midrule
 All & 3396 & \textbf{0.0000} & \textbf{0.909} & \textbf{0.721} & \textbf{0.293} & \textbf{0.136} \\
\midrule
\multicolumn{7}{l}{\emph{All parts, published results~\citep{cadena}}} \\
CAD-Recode & 3396 & 0.2790 & 0.686 & 0.489 & 1.061 & 0.905 \\
cadrille & 3396 & \underline{0.0080} & 0.688 & 0.498 & 1.267 & 1.125 \\
CADReasoner & 3396 & 0.3990 & 0.683 & 0.529 & 1.971 & 1.833 \\
CADEvolve & 3396 & 0.1080 & 0.730 & 0.529 & 1.312 & 1.143 \\
CADFit$^\dagger$ & 3396 & 0.1880 & 0.757 & 0.548 & 0.713 & 0.536 \\
CADENA-SFT greedy & 3396 & 0.1280 & 0.739 & 0.580 & 0.964 & 0.774 \\
CADENA-RL greedy & 3396 & 0.0160 & \underline{0.876} & \underline{0.670} & \underline{0.331} & \underline{0.176} \\
 \bottomrule
 \end{tabular}
 \end{table}

Table~\ref{tab:comparison} compares CADFather with the published CADENA results and the baselines evaluated in that paper. On DeepCAD, CADFather reaches a mean IoU of 0.987 against 0.966 for CADENA-RL with sampling; on Fusion360, 0.976 against 0.952; on MCB, 0.913 against 0.895. The invalidity ratio is zero on all three datasets: every part ends with a closed solid. The IoU gain over sampling is of similar size on all three datasets (0.018--0.024), and mean GMS is higher as well (0.983, 0.960, and 0.766 against 0.974, 0.942, and 0.751). Median CD is lower on all three datasets at both sampling densities (at 30000 points, 0.039, 0.034, and 0.083 against 0.042, 0.036, and 0.089). The stepwise tool uses the same CADENA-RL checkpoint, so the gain does not come from a stronger generator. The ablations (Section~\ref{sec:ablations}) compare configurations of CADFather with each other rather than with CADENA; among them, removing deterministic proposals costs the most: it lowers the mean score by 0.012--0.110 on the 1000-part subsets and raises IR from zero to 0.007--0.124, whereas removing the optimizer lowers the mean score by only 0.004--0.006 and leaves IR at zero. Mean GMS on MCB remains 0.766, well below the IoU; the two means cover different subsets of targets, but the gap suggests that matching volume does not imply matching surface structure on mechanical components. Across DeepCAD, Fusion360, and MCB, CADFather achieves the best reported IR, mean IoU, mean GMS, and median CD at both sampling densities among all methods compared in Table~\ref{tab:comparison}.

\subsection{Evaluation on CADENA-Bench}
\label{sec:cadenabench}

Table~\ref{tab:cadenabench} reports CADFather on CADENA-Bench with the configuration of the main runs. The invalidity ratio is again zero, and the mean IoU of 0.909 over the 2240 targets with defined IoU is close to that on MCB (0.913). As on MCB, where 58\% of the targets have a defined IoU, only 66\% do here, against 95--99\% on DeepCAD and Fusion360, so on mechanical parts GMS, which covers all targets, carries a larger part of the comparison. The shaft-and-bushing and lever-and-profile families are reconstructed best; among families with enough parts to compare, the gear-and-bearing and spring-and-fastener families have the lowest IoU and GMS. Over all parts, the mean IoU of 0.909 and GMS of 0.721 compare with 0.876 and 0.670 for CADENA-RL greedy, the strongest published row; CADENA does not report its sampling variant on this benchmark. The three sources reach the same mean IoU (0.909--0.910 for CADNET, MCB, and TMCAD).

\subsection{CADBench and BenchCAD}
\label{sec:newbench}

\begin{table}[!tbp]
\centering
\begin{minipage}[t]{0.44\linewidth}
 \centering
\setlength{\tabcolsep}{3.5pt}
\caption{CADFather on CADBench and BenchCAD over all parts, under our protocol (Section~\ref{sec:protocol}). IoU and GMS are means, CD values are medians scaled by $10^3$.}
\label{tab:newbench}
\small
\begin{tabular}{lrrrrr}
\toprule
Benchmark & IR & IoU & GMS & CD$_{8k}$ & CD$_{30k}$ \\
\midrule
CADBench & 0.0007 & 0.904 & 0.815 & 0.145 & 0.048 \\
BenchCAD & 0.0000 & 0.948 & 0.873 & 0.160 & 0.050 \\
\bottomrule
\end{tabular}
\end{minipage}\hfill
\begin{minipage}[t]{0.52\linewidth}
\centering
\caption{Mesh-to-CAD results on CADBench with its evaluation code, aggregated over all 18000 parts. Rows other than CADFather are published results of~\citet{cadbench} (clean meshes; mean of three runs for non-deterministic methods). CADFather is a single run; its programs use our operation library on top of CadQuery, so program-length metrics (token and operation counts) are not comparable and are omitted. CD is scaled as in~\citet{cadbench}.}
\label{tab:cadbench_official}
\small
\begin{tabular}{lrrrr}
\toprule
Method & IoU $\uparrow$ & SIoU $\uparrow$ & CD $\downarrow$ & VSR $\uparrow$ \\
\midrule
CAD-Recode & 0.673 & 0.503 & 0.056 & 0.912 \\
CADEvolve & 0.707 & 0.498 & 0.062 & 0.967 \\
cadrille & 0.687 & 0.517 & 0.056 & 0.939 \\
CADFit & \underline{0.859} & \underline{0.679} & \underline{0.038} & \textbf{0.981} \\
CADFather & \textbf{0.930} & \textbf{0.787} & \textbf{0.029} & \underline{0.968} \\
\bottomrule
\end{tabular}
\end{minipage}
\end{table}

Table~\ref{tab:newbench} reports CADFather on CADBench and BenchCAD under our protocol. On CADBench, 12 of the 18000 parts end without a valid prediction (IR 0.0007). On all of them the target mesh is so dense that computing the IoU of an overlapping candidate exceeds the 30\,s execution limit, so every candidate fails. The mean IoU over the 17314 targets with defined IoU is 0.904. Quality follows the families: the Base family of sketch-and-extrude parts is reconstructed almost exactly (IoU 0.980, GMS 0.974), while the Organic family of object-like meshes is the hardest (IoU 0.759, GMS 0.548). On BenchCAD, every prediction is valid, the mean IoU is 0.948 over 16350 targets, and IoU and GMS decrease monotonically from the easy to the hard tier (0.956 to 0.938 and 0.886 to 0.859). The lowest scores fall on helical springs and twist drills, which remain poor even on targets with defined IoU.

For comparison with published methods, Table~\ref{tab:cadbench_official} evaluates the final programs of all 18000 CADBench parts with the benchmark's own evaluation code and aggregation~\citep{cadbench}: aligned IoU and surface IoU (SIoU) are medians with failures counted as zero, CD is the median over successfully evaluated programs, and the valid shape rate (VSR) is the fraction of programs that produce a valid shape; each metric is aggregated over the 14 complexity splits weighted by their size. CADFather reaches higher IoU and SIoU and lower CD than all published mesh-to-CAD methods, including CADFit, and is better than CADFit on each of the 14 splits in these three metrics. Its VSR is lower than that of CADFit: 576 parts exceed the evaluator's 30\,s limit, which covers program execution and all metric computations for a part.

\begin{table}[!tbp]
\centering
\caption{Vision2Code voxel IoU and invalid rate on BenchCAD. Image-conditioned methods receive four orthographic views; with tools, the model also gets a Python sandbox to render, measure, and iterate before submitting. Frontier models are taken from the BenchCAD leaderboard~\citep{benchcadboard}; $^*$~self-reported by the vendor, without IR. CADENA-RL is reported by~\citet{cadena}, the other open methods by~\citet{benchcad}. Parts: the number of the 17\,900 BenchCAD parts a row is evaluated on; 1000 is a random subset drawn by the vendor (for Claude, a mean over five runs) and by us for CADFather; n/s, not stated. Other rows are published by~\citet{benchcad} and~\citet{cadena}.}
\label{tab:benchcad_board}
\small
\begin{tabular}{llrrr}
\toprule
Method & Input & Parts & Voxel IoU $\uparrow$ & IR (\%) $\downarrow$ \\
\midrule
Claude Sonnet 5.5 (max)$^*$ & image & 1000 & 0.747 & -- \\
Claude Opus 5.5 (max)$^*$ & image & 1000 & 0.730 & -- \\
GPT-5.6 Sol (max)$^*$ & image & n/s & 0.706 & -- \\
Gemini 3.1 Pro (thinking) & image & n/s & 0.355 & 18.5 \\
Claude Opus 4.7 (thinking) & image & n/s & 0.279 & 3.5 \\
\midrule
Claude Sonnet 5.5 (max)$^*$ & image + tools & 1000 & \underline{0.963} & -- \\
Claude Opus 5.5 (max)$^*$ & image + tools & 1000 & 0.962 & -- \\
GPT-6 Astra$^*$ & image + tools & 1000 & 0.959 & -- \\
GPT-5.6 Sol (max)$^*$ & image + tools & n/s & 0.834 & -- \\
\midrule
cadrille-RL & image & n/s & 0.068 & 8.8 \\
CADEvolve v3 & image & n/s & 0.750 & 7.3 \\
Qwen3-VL-2B RL, in-distribution~\citep{benchcad} & image & n/s & 0.752 & 1.1 \\
\midrule
CADENA-RL & mesh & 17\,900 & 0.910 & \underline{0.9} \\
CADFather & mesh & 1000 & \textbf{0.968} & \textbf{0.0} \\
\bottomrule
\end{tabular}
\end{table}

\begin{figure}[!tbp]
\centering
\includegraphics[width=0.6\linewidth]{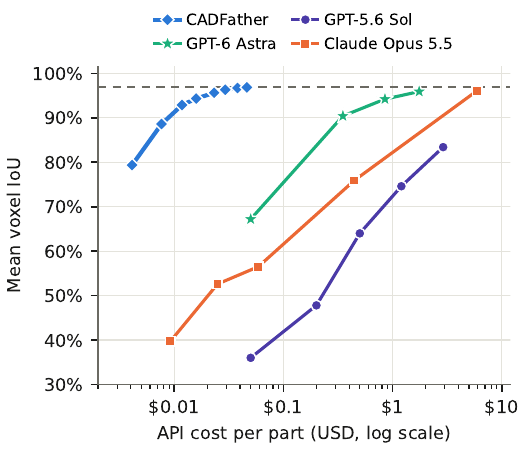}
\caption{Voxel IoU on BenchCAD against API cost per part (log scale). CADFather is evaluated on 1000 parts drawn at random; each point is the mean after the first $k$ assistant steps of one full-budget run ($k = 1, 2, 3, 4, 6, 8, 12, 24$), and the dashed line marks its full-budget level. Assistant and generator tokens are both priced at the OpenRouter rate for Qwen3.8-27B (\$0.425 and \$2.55 per million input and output tokens). Frontier curves are taken from the BenchCAD leaderboard~\citep{benchcadboard}.}
\label{fig:benchcad_cost}
\end{figure}

Published BenchCAD results follow its Vision2Code protocol: voxel IoU on a $64^3$ grid and the invalid rate (IR), i.e., the fraction of generated programs that fail to execute~\citep{benchcad}. Table~\ref{tab:benchcad_board} places CADFather beside the published entries; we score it with the voxelization from the BenchCAD evaluation code. CADFather reaches a voxel IoU of 0.968 with no invalid predictions, against 0.910 for CADENA-RL, the other mesh-conditioned entry, and at most 0.752 for image-conditioned systems without tools. Frontier models with a Python sandbox self-report up to 0.963 on 1\,000-part subsets.

Figure~\ref{fig:benchcad_cost} relates voxel IoU to API cost per part. CADFather reaches an IoU of 0.943 after four assistant steps at \$0.016 per part, and 0.968 at its full budget of \$0.046 per part. For comparison, Claude Opus~5.5 and GPT-6~Astra with tools report 0.962 at \$5.94 per part and 0.959 at \$1.75 per part, respectively.

\FloatBarrier

\subsection{Ablations}
\label{sec:ablations}
The ablations separate the contribution of each tool family from the effect of additional compute. Each arm removes tools from the full configuration and keeps all budgets, limits, and model servers identical.
\begin{itemize}
\item \emph{Stepwise only}: deterministic proposals and optimization are removed; the generator-sample budget is unchanged, so the arm tests whether tool diversity adds value beyond sampling at a matched generation budget. This arm is not CADENA sampling: CADFather never extends an invalid candidate, so a part whose first operations all produce non-watertight meshes ends invalid, whereas CADENA's sampling continues from any candidate that receives a reward.
\item \emph{w/o deterministic proposals}: both entry points of the deterministic tool are removed.
\item \emph{w/o optimization}: the parameter optimizer is removed.
\end{itemize}
Differences are reported against the full-tool control on the same subset as paired per-part differences in the score of Eq.~\ref{eq:score}, with 95\% bootstrap confidence intervals over parts (2000 resamples). Unlike the IoU and GMS means, the score difference covers all parts, with an invalid prediction scored as zero.
 
\begin{table}[!tbp]
\centering
\caption{Ablations on 1000-part subsets. $\Delta s$ is the paired difference in mean score (Eq.~\ref{eq:score}) to the full-tool control with a 95\% bootstrap interval, taken over all 1000 parts with invalid predictions scored as zero; IR is a fraction. IoU and GMS are means over valid predictions; IoU covers only the 565 (MCB), 983 (DeepCAD), and 949 (Fusion360) targets with defined IoU.}
\label{tab:ablations}
\small
\setlength{\tabcolsep}{4pt}
\begin{tabular}{llrrrr}
 \toprule
Dataset & Configuration & IR & IoU & GMS & $\Delta s$ [95\% CI] \\
\midrule
MCB & Full (control) & \textbf{0.000} & \textbf{0.904} & \textbf{0.756} & --- \\
 & Stepwise only & \underline{0.123} & 0.872 & 0.735 & $-0.1135$ [$-0.1303$, $-0.0979$] \\
 & w/o deterministic proposals & 0.124 & 0.882 & 0.739 & $\underline{-0.1095}$ [$-0.1265$, $-0.0942$] \\
 & w/o optimization & \textbf{0.000} & \underline{0.895} & \underline{0.750} & $\mathbf{-0.0061}$ [$-0.0088$, $-0.0034$] \\
 \midrule
DeepCAD & Full (control) & \textbf{0.000} & \textbf{0.984} & \textbf{0.983} & --- \\
 & Stepwise only & \underline{0.007} & 0.960 & \underline{0.979} & $-0.0201$ [$-0.0254$, $-0.0152$] \\
 & w/o deterministic proposals & \underline{0.007} & \underline{0.976} & \underline{0.979} & $\underline{-0.0124}$ [$-0.0176$, $-0.0076$] \\
 & w/o optimization & \textbf{0.000} & 0.975 & \textbf{0.983} & $\mathbf{-0.0045}$ [$-0.0059$, $-0.0030$] \\
\midrule
Fusion360 & Full (control) & \textbf{0.000} & \textbf{0.976} & \textbf{0.958} & --- \\
 & Stepwise only & \underline{0.012} & 0.952 & 0.951 & $-0.0266$ [$-0.0335$, $-0.0200$] \\
 & w/o deterministic proposals & \underline{0.012} & 0.966 & \underline{0.952} & $\underline{-0.0192}$ [$-0.0260$, $-0.0129$] \\
 & w/o optimization & \textbf{0.000} & \underline{0.968} & \textbf{0.958} & $\mathbf{-0.0042}$ [$-0.0062$, $-0.0023$] \\
\bottomrule
\end{tabular}
\end{table}

\begin{figure}[!tbp]
\centering
\includegraphics[width=0.85\linewidth]{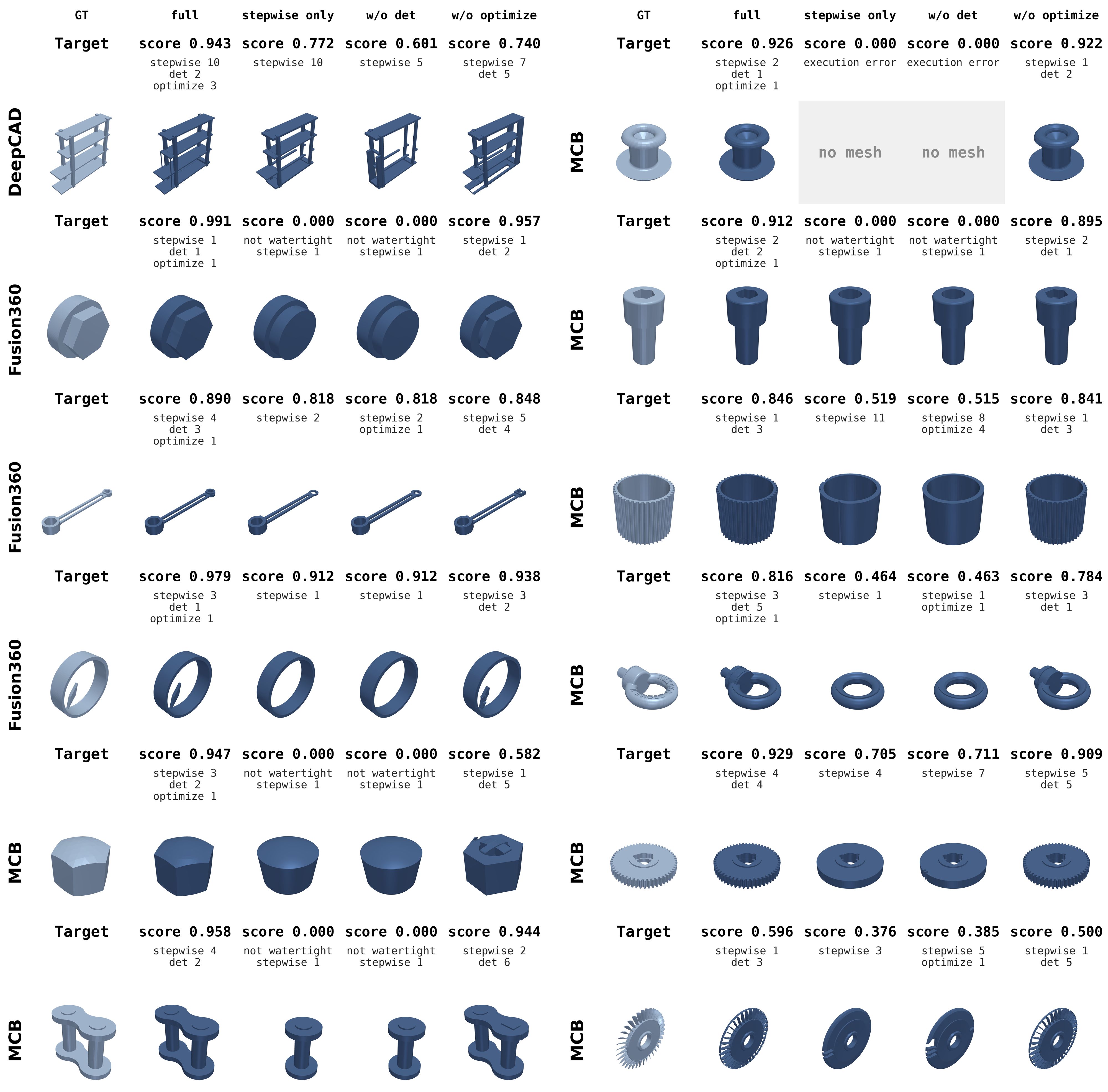}
\caption{Examples from the ablation subsets. Each row is one part: the target (GT) and the final results of the full configuration, stepwise only, w/o deterministic proposals (w/o det), and w/o optimization, rendered in a common frame. Above each result are its score (Eq.~\ref{eq:score}; zero for an invalid prediction) and the number of tool calls on the lineage of that result (det counts both deterministic entry points).}

\label{fig:ablation_examples}
\end{figure}

Removing all non-generative tools changes the mean score by $-0.114$ on MCB, about as much as removing deterministic proposals alone ($-0.110$) and far more than removing optimization alone ($-0.006$). Deterministic proposals carry most of the gain: without them the invalidity ratio on MCB rises from zero to 0.12. Nearly all of these failures are non-watertight results of parts on which no sampled first operation yields a closed solid, so the search cannot proceed from a valid base; deterministic proposals supply such a base. Since an invalid prediction scores zero, this rise alone accounts for most of the $-0.11$ on MCB, while the IoU and GMS of the remaining valid predictions drop by only 0.02--0.03. On MCB, the optimizer adds almost nothing on top of stepwise generation, and \emph{w/o deterministic proposals} and \emph{stepwise only} are close. On DeepCAD and Fusion360 the ordering of the arms is the same at a smaller scale, but there the optimizer on top of stepwise generation adds about 0.008, more than its own contribution in the full configuration. Candidate-level best updates, not tool invocation counts, indicate whether a tool supplied a useful result: on the three full test sets, 27--36\% of stepwise calls, 25--29\% of deterministic-proposal calls, and 18--34\% of optimizer calls raised the protected best. These shares favor tools that are called early in a part, when the best is still easy to raise. Figure~\ref{fig:ablation_examples} shows parts on which the full configuration scores above all three ablations: in these examples, the arms without deterministic proposals lose a whole feature, produce a non-watertight body, or fail to execute.

\subsection{Efficiency and Qualitative Analysis}

\begin{table}[!tbp]
\centering
\caption{Per-part cost of CADFather on the full test sets and CADENA-Bench (means). Tool calls count requests; generator samples count proposals across stepwise requests. A part is finished by the assistant when it ends with a \texttt{finish} call while an affordable action remains.}
\label{tab:efficiency}
\small
\begin{tabular}{lrrrr}
\toprule
 & DeepCAD & Fusion360 & MCB & CADENA-Bench \\
\midrule
Assistant steps & 6.37 & 8.41 & 12.25 & 13.04 \\
Tool calls & 8.67 & 12.43 & 20.57 & 23.01 \\
\quad stepwise / deterministic / optimizer & 4.57 / 1.81 / 2.29 & 6.60 / 2.72 / 3.10 & 10.96 / 4.94 / 4.68 & 12.38 / 5.34 / 5.29 \\
Generator samples & 19.0 & 31.9 & 61.1 & 67.1 \\
Calls yielding an executable candidate (\%) & 83.3 & 85.9 & 89.8 & 90.0 \\
Calls raising the protected best (\%) & 29.0 & 29.6 & 30.0 & 36.1 \\
Parts finished by the assistant (\%) & 86.1 & 73.4 & 37.3 & 25.0 \\
Wall time per part (s) & 82.6 & 119.9 & 245.6 & 306.7 \\
\bottomrule
\end{tabular}
\end{table}

Table~\ref{tab:efficiency} reports the per-part cost of the full configuration. Wall time depends on hardware, model serving, and concurrency; it is reported for the stated deployment and does not by itself establish an algorithmic speedup.

The assistant ends 86.1\%, 73.4\%, and 37.3\% of parts itself on DeepCAD, Fusion360, and MCB, and 90.2\%, 82.3\%, and 50.9\% of these endings give the reason \texttt{shape}; the remaining parts stop when no affordable action is left or at the step, execution, or wall-time limits. Mechanical parts take about twice as many assistant steps as DeepCAD parts (12.3 on MCB against 6.4 on DeepCAD and 8.4 on Fusion360) and about three times the wall time. CADENA-Bench is the most expensive set: 13.0 steps and 307\,s per part, with only 25.0\% of parts ended by the assistant (55.2\% of them with \texttt{shape}). Figure~\ref{fig:trajectory} shows a trace whose decisions and measurements were verified against the logs. 

\begin{figure}[!tbp]
\centering
\includegraphics[width=\linewidth]{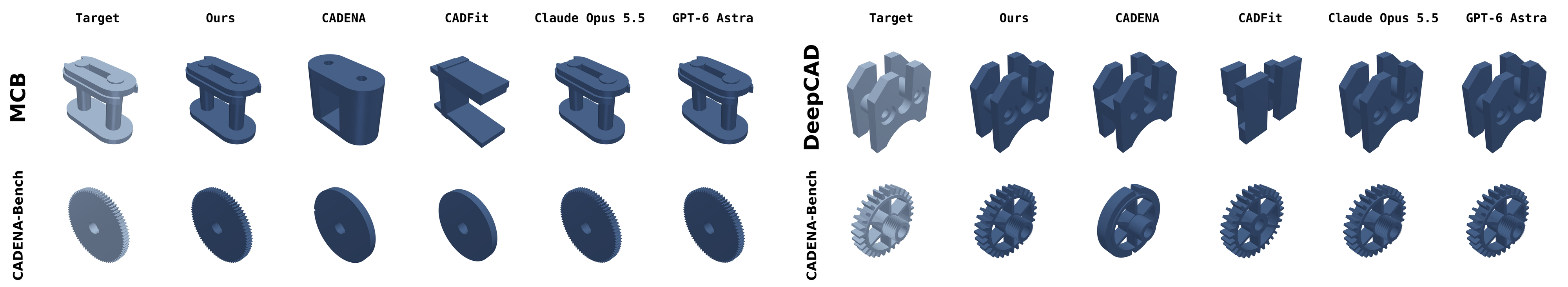}
\caption{Comparison with CADENA, CADFit, and two frontier coding agents. Each row is one part: the target, CADFather (Ours), CADENA-RL, CADFit$^\dagger$, and the reconstructions of Claude Opus 5.5 and GPT-6 Astra. The frontier agents wrote CadQuery code with access to the target mesh and received the score against the target after every attempt.}
\label{fig:frontier_examples}
\end{figure}

Figure~\ref{fig:frontier_examples} shows parts on which CADFather is well ahead of CADENA and CADFit. These methods miss whole features, such as the pins of a chain link or the teeth of a gear, or build a different overall layout, while CADFather recovers the structure of the part. Two frontier coding agents, Claude Opus 5.5 and GPT-6 Astra, given the target mesh and the score after every attempt, reach scores close to those of CADFather on these parts.

\begin{figure}[!h]
\centering
\includegraphics[width=0.7\linewidth]{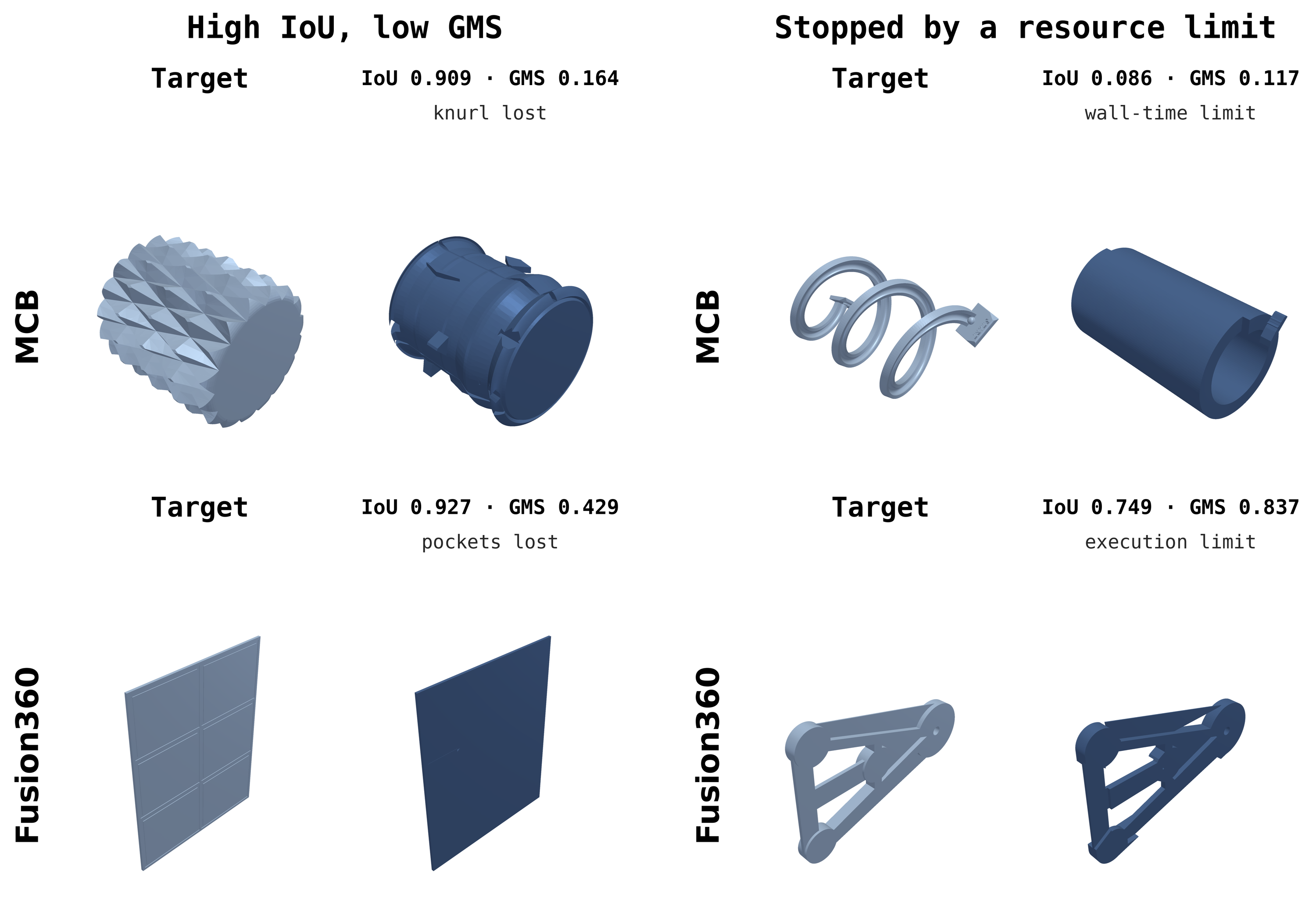}
\caption{Failure cases from the full test sets. Each pair shows the target and the CADFather result in a common frame, with IoU and GMS above the result. Left: high IoU with low GMS, where the volume is matched but the knurl (MCB) and the shallow pockets (Fusion360) are missing. Right: parts stopped by the wall-time limit (MCB, a spring reconstructed as a tube) and by the execution limit (Fusion360, a bracket with incomplete ribs).}
\label{fig:failure_examples}
\end{figure}

Typical failure cases are an incorrect initial topology that later operations cannot recover, an attractive overlap score with poor surface agreement, and a part terminated by a resource limit. In the first case, the branches that survive all descend from one wrong root: a solid block is rebuilt as a hollow frame, or a ladder's rungs merge into one plate. In the second case (Figure~\ref{fig:failure_examples}, left), the volume is matched while fine surface structure such as a knurl or shallow pockets is never built; on the panel, the assistant judged the remaining gap numerical and stopped. A resource limit can stop a part that was still improving, as on the bracket, or exhaust the wall time on a complex geometry before a correct base exists, as on the helical spring rebuilt as a tube (Figure~\ref{fig:failure_examples}, right).

\FloatBarrier

\section{Limitations}
CADFather remains limited by the operations supported by its CAD representation and the proposals its tools can produce. Parameter optimization can refine an existing program but cannot correct an unsuitable sequence of operations; likewise, retaining earlier candidates does not help if a useful alternative was never generated. Search limits may also terminate a trajectory that would have succeeded with more computation.

The assistant decides which candidates and tools to use from incomplete visual and numerical evidence. Its decisions can depend on the prompt, model behavior, and which candidates are available for inspection. This flexibility therefore does not guarantee that the best reconstruction will be found.

Our evaluation measures geometric reconstruction and program validity, not recovery of the original design history, manufacturing intent, or engineering constraints. It covers individual parts rather than assemblies, tolerances, or simulation-validated designs. CADFather also incurs computational costs for planning, inspection, and executing alternatives. The ablations use fixed budgets but are limited to 1,000-part subsets; adapting agent baselines such as CAD-Assistant and IterCAD for a direct comparison remains future work.

\section{Conclusion}
We presented CADFather, an autonomous agent for reconstructing executable CAD programs from target meshes. A vision-language assistant inspects the target and intermediate reconstructions, selects candidate programs and tools, and uses execution feedback to guide subsequent decisions. Stepwise generation and deterministic proposals provide alternative modeling operations, while parameter optimization refines existing programs. The system preserves the best valid candidate as it explores alternative reconstructions, without additional training of either model.

Across the full DeepCAD, Fusion360, and MCB test sets, CADFather improves mean IoU and GMS and reduces median Chamfer distance relative to the published CADENA-RL sampling results. It produces valid outputs for every part in these tests under its stricter validity rule. CADFather also improves on the published CADENA-RL greedy results on CADENA-Bench. On CADBench, it achieves higher IoU and surface IoU and lower Chamfer distance than the reported mesh-to-CAD baselines, although its valid shape rate is below that of CADFit. On BenchCAD, CADFather achieves higher voxel IoU than the frontier agents shown, at up to 100× lower estimated inference cost. Overall, the results support coordinating complementary reconstruction tools rather than committing to a single source of CAD operations. Direct comparisons with other CAD agents and a controlled evaluation of the assistant's tool-selection policy remain future work.

\bibliographystyle{unsrtnat}
\bibliography{references}

\begin{thebibliography}{36}
\providecommand{\natexlab}[1]{#1}
\providecommand{\url}[1]{\texttt{#1}}
\expandafter\ifx\csname urlstyle\endcsname\relax
  \providecommand{\doi}[1]{doi: #1}\else
  \providecommand{\doi}{doi: \begingroup \urlstyle{rm}\Url}\fi

\bibitem[Rukhovich et~al.(2025)Rukhovich, Dupont, Mallis, Cherenkova, Kacem,
  and Aouada]{cadrecode}
Danila Rukhovich, Elona Dupont, Dimitrios Mallis, Kseniya Cherenkova, Anis
  Kacem, and Djamila Aouada.
\newblock {CAD-Recode}: Reverse engineering {CAD} code from point clouds.
\newblock In \emph{IEEE/CVF International Conference on Computer Vision
  (ICCV)}, 2025.

\bibitem[Kolodiazhnyi et~al.(2025)Kolodiazhnyi, Tarasov, Zhemchuzhnikov,
  Nikulin, Zisman, Vorontsova, Konushin, Kurenkov, and Rukhovich]{cadrille}
Maksim Kolodiazhnyi, Denis Tarasov, Dmitrii Zhemchuzhnikov, Alexander Nikulin,
  Ilya Zisman, Anna Vorontsova, Anton Konushin, Vladislav Kurenkov, and Danila
  Rukhovich.
\newblock cadrille: Multi-modal {CAD} reconstruction with reinforcement
  learning.
\newblock \emph{arXiv preprint arXiv:2505.22914}, 2025.

\bibitem[Elistratov et~al.(2026)Elistratov, Barannikov, Ivanov, Khrulkov,
  Konushin, Kuznetsov, and Zhemchuzhnikov]{cadevolve}
Maksim Elistratov, Marina Barannikov, Gregory Ivanov, Valentin Khrulkov, Anton
  Konushin, Andrey Kuznetsov, and Dmitrii Zhemchuzhnikov.
\newblock {CADEvolve}: Creating realistic {CAD} via program evolution.
\newblock \emph{arXiv preprint arXiv:2602.16317}, 2026.

\bibitem[Kabisov et~al.(2026{\natexlab{a}})Kabisov, Kirichuk, Volkov, Savrasov,
  Barannikov, Konushin, Kuznetsov, and Zhemchuzhnikov]{cadreasoner}
Soslan Kabisov, Vsevolod Kirichuk, Andrey Volkov, Gennadii Savrasov, Marina
  Barannikov, Anton Konushin, Andrey Kuznetsov, and Dmitrii Zhemchuzhnikov.
\newblock {CADReasoner}: Iterative program editing for {CAD} reverse
  engineering.
\newblock In \emph{IEEE/CVF Conference on Computer Vision and Pattern
  Recognition (CVPR), Findings}, 2026{\natexlab{a}}.

\bibitem[Kabisov et~al.(2026{\natexlab{b}})Kabisov, Savrasov, Elistratov,
  Rodriguez, Ignatiev, Gavrilov, Uzdenov, Boyko, Pasechnik, Konushin,
  Kuznetsov, and Zhemchuzhnikov]{cadena}
Soslan Kabisov, Gennadiy Savrasov, Maksim Elistratov, Antonio Rodriguez, Daniil
  Ignatiev, Nikita Gavrilov, Rustam Uzdenov, Alexey~I. Boyko, Igor Pasechnik,
  Anton Konushin, Andrey Kuznetsov, and Dmitrii Zhemchuzhnikov.
\newblock {CADENA}: Stepwise {CAD} reverse engineering.
\newblock \emph{arXiv preprint arXiv:2608.00799}, 2026{\natexlab{b}}.
\newblock \doi{10.48550/arXiv.2608.00799}.
\newblock URL \url{https://arxiv.org/abs/2608.00799}.

\bibitem[Nehme et~al.(2026)Nehme, Whalen, and Ahmed]{cadfit}
Ghadi Nehme, Eamon Whalen, and Faez Ahmed.
\newblock {CADFit}: Precise mesh-to-{CAD} program generation with hybrid
  optimization.
\newblock In \emph{International Conference on Machine Learning (ICML)}, 2026.

\bibitem[Cascaval et~al.(2022)Cascaval, Shalah, Quinn, Bodik, Agrawala, and
  Schulz]{differentiablecad}
Dan Cascaval, Mira Shalah, Phillip Quinn, Rastislav Bodik, Maneesh Agrawala,
  and Adriana Schulz.
\newblock Differentiable 3d {CAD} programs for bidirectional editing.
\newblock \emph{Computer Graphics Forum}, 41\penalty0 (2):\penalty0 309--323,
  2022.
\newblock \doi{10.1111/cgf.14476}.
\newblock URL \url{https://arxiv.org/abs/2110.01182}.

\bibitem[Yuan et~al.(2024)Yuan, Bousseau, Pan, Zhang, Mitra, and Li]{diffcsg}
Haocheng Yuan, Adrien Bousseau, Hao Pan, Chengquan Zhang, Niloy~J. Mitra, and
  Changjian Li.
\newblock {DiffCSG}: Differentiable {CSG} via rasterization.
\newblock In \emph{SIGGRAPH Asia 2024 Conference Papers}, pages 1--10, 2024.
\newblock \doi{10.1145/3680528.3687608}.
\newblock URL \url{https://arxiv.org/abs/2409.01421}.

\bibitem[Mallis et~al.(2025)Mallis, Karadeniz, Cavada, Rukhovich,
  Foteinopoulou, Cherenkova, Kacem, and Aouada]{cadassistant}
Dimitrios Mallis, Ahmet~Serdar Karadeniz, Sebastian Cavada, Danila Rukhovich,
  Niki Foteinopoulou, Kseniya Cherenkova, Anis Kacem, and Djamila Aouada.
\newblock {CAD-Assistant}: Tool-augmented {VLLMs} as generic {CAD} task
  solvers.
\newblock In \emph{IEEE/CVF International Conference on Computer Vision
  (ICCV)}, 2025.

\bibitem[Hu et~al.(2026)Hu, Ai, Wen, Li, Zou, Li, Deng, Cai, Zhou, Cai, Fu,
  Yang, Zhang, Shi, and Yang]{itercad}
Tao Hu, Jiaxin Ai, Licheng Wen, Xueheng Li, Shu Zou, Siqi Li, Nianchen Deng,
  Xinyu Cai, Hongbin Zhou, Pinlong Cai, Daocheng Fu, Yu~Yang, Hairong Zhang,
  Botian Shi, and Xuemeng Yang.
\newblock {IterCAD}: An iterative multimodal agent for visually-grounded {CAD}
  generation and editing.
\newblock \emph{arXiv preprint arXiv:2606.13368}, 2026.

\bibitem[Wu et~al.(2021)Wu, Xiao, and Zheng]{deepcad}
Rundi Wu, Chang Xiao, and Changxi Zheng.
\newblock {DeepCAD}: A deep generative network for computer-aided design
  models.
\newblock In \emph{IEEE/CVF International Conference on Computer Vision
  (ICCV)}, 2021.

\bibitem[Willis et~al.(2021)Willis, Pu, Luo, Chu, Du, Lambourne, Solar-Lezama,
  and Matusik]{fusion360}
Karl D.~D. Willis, Yewen Pu, Jieliang Luo, Hang Chu, Tao Du, Joseph~G.
  Lambourne, Armando Solar-Lezama, and Wojciech Matusik.
\newblock {Fusion 360 Gallery}: A dataset and environment for programmatic
  {CAD} construction from human design sequences.
\newblock \emph{ACM Transactions on Graphics}, 40\penalty0 (4), 2021.

\bibitem[Xu et~al.(2022)Xu, Willis, Lambourne, Cheng, Jayaraman, and
  Furukawa]{skexgen}
Xiang Xu, Karl D.~D. Willis, Joseph~G. Lambourne, Chin-Yi Cheng, Pradeep~Kumar
  Jayaraman, and Yasutaka Furukawa.
\newblock {SkexGen}: Autoregressive generation of {CAD} construction sequences
  with disentangled codebooks.
\newblock In \emph{International Conference on Machine Learning (ICML)}, 2022.

\bibitem[Para et~al.(2021)Para, Bhat, Guerrero, Kelly, Mitra, Guibas, and
  Wonka]{sketchgen}
Wamiq~Reyaz Para, Shariq~Farooq Bhat, Paul Guerrero, Tom Kelly, Niloy Mitra,
  Leonidas Guibas, and Peter Wonka.
\newblock {SketchGen}: Generating constrained {CAD} sketches.
\newblock In \emph{Advances in Neural Information Processing Systems
  (NeurIPS)}, 2021.

\bibitem[Seff et~al.(2022)Seff, Zhou, Richardson, and Adams]{vitruvion}
Ari Seff, Wenda Zhou, Nick Richardson, and Ryan~P. Adams.
\newblock {Vitruvion}: A generative model of parametric {CAD} sketches.
\newblock In \emph{International Conference on Learning Representations
  (ICLR)}, 2022.

\bibitem[Wu et~al.(2024)Wu, Khasahmadi, Katz, Jayaraman, Pu, Willis, and
  Liu]{cadvlm}
Sifan Wu, Amir Khasahmadi, Mor Katz, Pradeep~Kumar Jayaraman, Yewen Pu, Karl
  Willis, and Bang Liu.
\newblock {CadVLM}: Bridging language and vision in the generation of
  parametric {CAD} sketches.
\newblock In \emph{European Conference on Computer Vision (ECCV)}, 2024.

\bibitem[Guo et~al.(2022)Guo, Liu, Pan, Liu, Tong, and Guo]{complexgen}
Haoxiang Guo, Shilin Liu, Hao Pan, Yang Liu, Xin Tong, and Baining Guo.
\newblock {ComplexGen}: {CAD} reconstruction by {B-Rep} chain complex
  generation.
\newblock \emph{ACM Transactions on Graphics}, 41\penalty0 (4), 2022.

\bibitem[Jayaraman et~al.(2023)Jayaraman, Lambourne, Desai, Willis, Sanghi, and
  Morris]{solidgen}
Pradeep~Kumar Jayaraman, Joseph~G. Lambourne, Nishkrit Desai, Karl D.~D.
  Willis, Aditya Sanghi, and Nigel J.~W. Morris.
\newblock {SolidGen}: An autoregressive model for direct {B-rep} synthesis.
\newblock \emph{Transactions on Machine Learning Research (TMLR)}, 2023.

\bibitem[Xu et~al.(2024)Xu, Lambourne, Jayaraman, Wang, Willis, and
  Furukawa]{brepgen}
Xiang Xu, Joseph~G. Lambourne, Pradeep~Kumar Jayaraman, Zhengqing Wang, Karl
  D.~D. Willis, and Yasutaka Furukawa.
\newblock {BrepGen}: A {B-rep} generative diffusion model with structured
  latent geometry.
\newblock \emph{ACM Transactions on Graphics}, 43\penalty0 (4), 2024.

\bibitem[Sharma et~al.(2022)Sharma, Goyal, Liu, Kalogerakis, and Maji]{csgnet}
Gopal Sharma, Rishabh Goyal, Difan Liu, Evangelos Kalogerakis, and Subhransu
  Maji.
\newblock Neural shape parsers for constructive solid geometry.
\newblock \emph{IEEE Transactions on Pattern Analysis and Machine
  Intelligence}, 2022.

\bibitem[Kania et~al.(2020)Kania, Zi{\k{e}}ba, and Kajdanowicz]{ucsgnet}
Kacper Kania, Maciej Zi{\k{e}}ba, and Tomasz Kajdanowicz.
\newblock {UCSG-Net}: Unsupervised discovering of constructive solid geometry
  tree.
\newblock In \emph{Advances in Neural Information Processing Systems
  (NeurIPS)}, 2020.

\bibitem[Yu et~al.(2022)Yu, Chen, Li, Sanghi, Shayani, Mahdavi-Amiri, and
  Zhang]{caprinet}
Fenggen Yu, Zhiqin Chen, Manyi Li, Aditya Sanghi, Hooman Shayani, Ali
  Mahdavi-Amiri, and Hao Zhang.
\newblock {CAPRI-Net}: Learning compact {CAD} shapes with adaptive primitive
  assembly.
\newblock In \emph{IEEE/CVF Conference on Computer Vision and Pattern
  Recognition (CVPR)}, 2022.

\bibitem[Uy et~al.(2022)Uy, Chang, Sung, Goel, Lambourne, Birdal, and
  Guibas]{point2cyl}
Mikaela~Angelina Uy, Yen-yu Chang, Minhyuk Sung, Purvi Goel, Joseph Lambourne,
  Tolga Birdal, and Leonidas Guibas.
\newblock {Point2Cyl}: Reverse engineering 3d objects from point clouds to
  extrusion cylinders.
\newblock In \emph{IEEE/CVF Conference on Computer Vision and Pattern
  Recognition (CVPR)}, 2022.

\bibitem[Ren et~al.(2022)Ren, Zheng, Cai, Li, and Zhang]{extrudenet}
Daxuan Ren, Jianmin Zheng, Jianfei Cai, Jiatong Li, and Junzhe Zhang.
\newblock {ExtrudeNet}: Unsupervised inverse sketch-and-extrude for shape
  parsing.
\newblock In \emph{European Conference on Computer Vision (ECCV)}, 2022.

\bibitem[Li et~al.(2023)Li, Guo, Zhang, and Yan]{secadnet}
Pu~Li, Jianwei Guo, Xiaopeng Zhang, and Dong-ming Yan.
\newblock {SECAD-Net}: Self-supervised {CAD} reconstruction by learning
  sketch-extrude operations.
\newblock In \emph{IEEE/CVF Conference on Computer Vision and Pattern
  Recognition (CVPR)}, 2023.

\bibitem[Doris et~al.(2025)Doris, Alam, Nobari, and Ahmed]{cadcodervlm}
Anna~C. Doris, Md~Ferdous Alam, Amin~Heyrani Nobari, and Faez Ahmed.
\newblock {CAD-Coder}: An open-source vision-language model for computer-aided
  design code generation.
\newblock \emph{arXiv preprint arXiv:2505.14646}, 2025.

\bibitem[Xu et~al.(2021)Xu, Peng, Cheng, Willis, and Ritchie]{zonegraphs}
Xianghao Xu, Wenzhe Peng, Chin-Yi Cheng, Karl D.~D. Willis, and Daniel Ritchie.
\newblock Inferring {CAD} modeling sequences using zone graphs.
\newblock In \emph{IEEE/CVF Conference on Computer Vision and Pattern
  Recognition (CVPR)}, 2021.

\bibitem[Feng et~al.(2026)Feng, Zhi, Zhang, Feng, and Zhao]{sovcad}
Zhaopeng Feng, Chen Zhi, Xuhong Zhang, Zhengwen Feng, and Xinkui Zhao.
\newblock {SOV-CAD}: Stepwise orthographic views guided {CAD} modeling sequence
  reconstruction.
\newblock In \emph{IEEE International Conference on Multimedia and Expo
  (ICME)}, 2026.

\bibitem[Gong et~al.(2026)Gong, Wu, Liu, and Tu]{toolcad}
Yifei Gong, Xing Wu, Wenda Liu, and Kang Tu.
\newblock {ToolCAD}: Exploring tool-using large language models in
  text-to-{CAD} generation with reinforcement learning.
\newblock In \emph{Annual Meeting of the Association for Computational
  Linguistics (ACL)}, 2026.

\bibitem[Deng et~al.(2026)Deng, Deng, Chen, Dai, Zhong, Li, Sun, Shi, and
  Huang]{cosmoagent}
Liyuan Deng, Shujian Deng, Yongkang Chen, Yongkang Dai, Zhihang Zhong, Linyang
  Li, Xiao Sun, Yilei Shi, and Huaxi Huang.
\newblock {COSMO-Agent}: Tool-augmented agent for closed-loop optimization,
  simulation, and modeling orchestration.
\newblock \emph{arXiv preprint arXiv:2604.05547}, 2026.

\bibitem[Kim et~al.(2020)Kim, Chi, Hu, Huang, and Ramani]{mcb}
Sangpil Kim, Hyung-Gun Chi, Xiao Hu, Qixing Huang, and Karthik Ramani.
\newblock A large-scale annotated mechanical components benchmark for
  classification and retrieval tasks with deep neural networks.
\newblock In \emph{European Conference on Computer Vision (ECCV)}, 2020.

\bibitem[Zou and Zhu(2025)]{tmcad}
Qiang Zou and Lizhen Zhu.
\newblock Bringing attention to cad: Boundary representation learning via
  transformer.
\newblock \emph{Computer-Aided Design}, page 103940, 2025.

\bibitem[Manda et~al.(2021)Manda, Bhaskare, and Muthuganapathy]{cadnet}
Bharadwaj Manda, Pranjal Bhaskare, and Ramanathan Muthuganapathy.
\newblock A convolutional neural network approach to the classification of
  engineering models.
\newblock \emph{arXiv preprint arXiv:2107.06481}, 2021.

\bibitem[Doris et~al.(2026)Doris, Sony, Nehme, Syla, Heyrani~Nobari, and
  Ahmed]{cadbench}
Anna~C. Doris, Jacob~Thomas Sony, Ghadi Nehme, Era Syla, Amin Heyrani~Nobari,
  and Faez Ahmed.
\newblock {CADBench}: A multimodal benchmark for {AI}-assisted {CAD} program
  generation.
\newblock \emph{arXiv preprint arXiv:2605.10873}, 2026.

\bibitem[Zhang et~al.(2026)Zhang, Liu, Chen, Li, Yang, Peng, and
  Chen]{benchcad}
Haozhe Zhang, Kaichen Liu, Miaomiao Chen, Lei Li, Shaojie Yang, Cheng Peng, and
  Hanjie Chen.
\newblock {BenchCAD}: A comprehensive, industry-standard benchmark for
  programmatic {CAD}.
\newblock \emph{arXiv preprint arXiv:2605.10865}, 2026.

\bibitem[{BenchCAD Team}(2026)]{benchcadboard}
{BenchCAD Team}.
\newblock {BenchCAD} leaderboard.
\newblock \url{https://benchcad.com/leaderboard}, 2026.
\newblock Accessed 2026-10-04.

\end{thebibliography}
\end{document}